\documentclass{article}
\usepackage{arxiv}
\usepackage[round, sort, numbers]{natbib}

\usepackage{lineno}
\usepackage{caption}
\usepackage{subcaption}
\usepackage{amssymb}
\usepackage{booktabs}
\usepackage{makecell}
\usepackage{xurl}
\usepackage[utf8]{inputenc} 
\usepackage[T1]{fontenc}    
\usepackage{hyperref}       
\usepackage{url}            
\usepackage{booktabs}       
\usepackage{amsfonts}       
\usepackage{nicefrac}       
\usepackage{microtype}      
\usepackage{lipsum}		
\usepackage{graphicx}
\usepackage{natbib}
\usepackage{doi}

\usepackage[justification=justified]{caption}

\title{ReCasNet: Improving consistency within the two-stage mitosis detection framework  }

\author{
  Chawan Piansaddhayanon$^{1,2}$ \And
  Sakun Santisukwongchote$^3$ \And
  Shanop Shuangshoti$^3$ \And
  Qingyi Tao$^4$ \And
  Sira Sriswasdi$^{2,5,6}$ \And
  Ekapol Chuangsuwanich$^{1,2,6}$ \And
  \vspace{-20pt} \\ 
  $^1$Department of Computer Engineering, Faculty of Engineering, Chulalongkorn University \\
  $^2$Chula Intelligent and Complex Systems, Faculty of Science, Chulalongkorn University \\
  $^3$Department of Pathology, King Chulalongkorn Memorial Hospital and Faculty of Medicine, Chulalongkorn University \\
  $^4$NVIDIA AI Technology Center, Singapore \\
  $^5$Computational Molecular Biology Group, Faculty of Medicine, Chulalongkorn University \\
  $^6$Center of Excellence in Computational Molecular Biology, Faculty of Medicine, Chulalongkorn University \\
  \texttt{\small schwan46494@gmail.com, boomskuun@gmail.com, shanop@gmail.com } \\
  \texttt{\small QTAO002@e.ntu.edu.sg, sira.sr@chula.ac.th, ekapol.c@chula.ac.th } }

\begin{document}
\maketitle

\begin{abstract}

Mitotic count (MC) is an important histological parameter for cancer diagnosis and grading, but the manual process for obtaining MC from whole-slide histopathological images is very time-consuming and prone to error. Therefore, deep learning models have been proposed to facilitate this process. Existing approaches utilize a two-stage pipeline: the detection stage for identifying the locations of potential mitotic cells and the classification stage for refining prediction confidences. However, this pipeline formulation can lead to inconsistencies in the classification stage due to the poor prediction quality of the detection stage and the mismatches in training data distributions between the two stages. In this study, we propose a Refine Cascade Network (ReCasNet), an enhanced deep learning pipeline that mitigates the aforementioned problems with three improvements. First, window relocation was used to reduce the number of poor quality false positives generated during the detection stage. Second, object re-cropping was performed with another deep learning model to adjust poorly centered objects. Third, improved data selection strategies were introduced during the classification stage to reduce the mismatches in training data distributions. 
ReCasNet was evaluated on two large-scale mitotic figure recognition datasets, canine cutaneous mast cell tumor (CCMCT) and canine mammary carcinoma (CMC), which resulted in up to 4.8\% percentage point improvements in the F1 scores for mitotic cell detection and 44.1\% reductions in mean absolute percentage error (MAPE) for MC prediction. Techniques that underlie ReCasNet can be generalized to other two-stage object detection networks and should contribute to improving the performances of deep learning models in broad digital pathology applications.

\end{abstract}

\keywords{
Mitotic count \and Whole slide image \and Object detection \and Image classification \and Multi-stage deep learning}

\section{Introduction}

Mitotic count is an important histologic parameter for cancer diagnosis and grading. Traditionally, mitotic count is obtained by manually counting mitotic figures through a light microscope. The hotspot area, usually spanning 10 high-power microscopic fields, that contain the highest density of mitotic figures in the whole histologic section(s) is identified and the number of mitotic figures in this area is reported as mitotic count. With the increasing use of digital pathology, whole slide image (WSI) is now routinely generated in several pathology laboratories. Nonetheless, mitotic count is still obtained by manual counting of mitotic figures on screen.  Conventionally, the manual process of obtaining MI is tedious and error-prone \cite {Veta}. Thus, several studies \cite{PAN2021107038} have utilized machine learning algorithms to assist pathologists by automatically recognizing mitotic figures in the WSI and proposing the hotspot area. Recently, deep learning has gained popularity due to its impressive image recognition performance compared to traditional approaches and is now widely used in a wide range of digital pathology applications, including histopathological image analysis \cite{Srinidhi_2021}.

Errors in mitotic figure detection by machine learning models can be attributed to the quality of data collection process and the ambiguity between mitotic figures in different cell division stages and other mitotic-like objects. First, each WSI is scanned on a single focal plane that could not be readjusted. As a result, many cells are out-of-focus and produce poor texture information. Additionally, the mitotic figures themselves can have diverse appearances across cell division stages and may be confused with other cell types or non-cell objects. Consequently, the classification of some mitotic figures could be highly subjective, which leads to drastically different mitotic counts reported by different experts \cite{Bertram2019ComputerizedCO}. Despite these problems, automated mitotic figure detection and mitotic count prediction is still considered as a crucial task in digital pathology and is an active area of research.

To develop models for automatic mitotic figure detection, datasets with expert annotations, such as the ICPR MITOS-2012 \cite {MITOS2012}, AMIDA 2013 \cite{AMIDA13}, ICPR MITOS-ATYPIA-2014 \cite{noauthor_mitos-atypia-14_nodate}, and TUPAC16 \cite{TUPAC16} challenges, can be used. However, as these datasets contain mitotic figure annotations only in the high power fields (HPF) corresponding to the hotspot ares, the model cannot fully learn from majority of the WSIs that were unannotated. Moreover, the number of the annotated mitotic figures in these datasets are low, often less than one thousand objects each. Recently, two large-scale mitotic figure datasets with annotations covering the whole slides have been released: the canine cutaneous mast cell tumor (CCMCT) \cite{CCMCT} dataset and the canine mammary carcinoma (CMC) \cite{CMC} dataset. The availability of these new datasets allow the model to learn from greatly increased mitotic figure and background diversity, which immediately improved the model's performance \cite{CCMCT}. Nonetheless, it should be noted that these datasets were annotated with fixed size circular bounding box with a radius of 25 pixels which do not perfectly capture the shape of mitotic figures and would lead to noises and errors during the training process.

Not only imperfection in data acquisition and annotation, but also the formulation of deep learning approach to solve the task plays an important role in the model's performance. Existing models for mitosis recognition often break the task into two stages: detection and classification \cite{Chen_Dou_Wang_Qin_Heng_2016, LI2018121, alom2019}. A main reason for this is because the sheer size of WSI prevents the model from operating directly on it. Instead, the WSI has to be broken down into smaller patches with a sliding window on which the inference is then performed to extract the locations of mitotic figures.
The detection stage proposes the locations of mitotic figures in the WSI by using deep object detection or segmentation models. After the mitotic figures are proposed, the classification stage then refines the prediction results by first extracting the position of each predicted mitotic figure and revising the corresponding image patch to make it center around the mitotic figure and to ensure that only one mitotic figure is contained within the patch. Each revised image patch is then fed to a deep object classifier to obtain a confidence score. The classification stage significantly improves the mitotic figure recognition performance because it overcomes the drawback of the detection stage which has to handle a much broader variety of image patches, some with no mitotic figure and some with multiple mitotic figures.

Despite the aforementioned benefits, a multi-stage pipeline also comes with a critical drawback; the classification stage suffers from inconsistency in the input data received from the detection stage and training distribution mismatch. As an inference is being performed at the detection stage, its outputs would inevitably consist of inaccurate object locations and poor quality bounding boxes, leading to inconsistently positioned objects at the image patch of the subsequent stage. 
The inconsistency results in classification stage performance degradation because most convolutional neural networks do not possess the shift-invariant property to properly handle the changes in distributions of object locations and bounding boxes produced by the detection stage \cite{Engstrom2017ARA}. The situation is further worsened with the use of a sliding window as it may split an object into pieces across multiple patches, which leads to additional poor-quality false positives. 
Inconsistency in training data distributions between the two stages is also non-negligible. While the detection stage learns the entire data distribution of the WSI, the classification stage mostly observes only mitotic figures and other similar-looking objects. This training distribution mismatch causes the classification stage to suffer from out-of-distribution problem when it receives inputs with no mitotic figure. DeepMitosis\cite{LI2018121} mitigates this problem by using all predictions, including low confidence ones, from the detection stage to train the classification stage. However, this method is impractical on large-scale datasets where hundreds of thousands of objects are proposed by the detector.

To address all of the aforementioned problems, we introduce Refine Cascade Network (ReCasNet), an enhanced deep learning pipeline to improve the recognition performance on large-scale mitotic figure recognition datasets. Our pipeline improves the performance of the classification stage by increasing the consistency of input data distribution and exposing the model to more informative data. First, we propose Window Relocation, a simple, effective method that overcomes the weakness of an overlapping sliding window by removing objects around the window border and re-evaluating them as the center of newly extracted patches. This method seeks to eliminate poor bounding boxes while requiring less computation cost than the overlapping sliding window. Second, we introduce an Object Center Adjustment Stage, a deep learning model responsible for bridging the gap between the classification stage and the detection stage. 
It generates new image patches that center on mitotic figures predicted by the detection stage and feed them to the classification stage to reduces the variance in input translation. Third, we improve the training data sampling process of the verification model (i.e., classification stage) of DeepMitosis by focusing on a subset of informative samples from the proposed objects on which the detector and the classifier disagree with each other the most.

We evaluated the performance of ReCasNet on the CCMCT and CMC datasets, two public large-scale datasets for mitotic figure assessment. ReCasNet achieved 83.2\% test F1 on the CCMCT dataset and 82.3\% test F1 on the CMC dataset, which correspond to +1.2 and +4.8 percentage point improvements over the baseline, respectively \cite{CCMCT, CMC}. An end-to-end evaluation on both datasets by comparing the HPF and mitotic count (MC) produced by ReCasNet to the ground truth annotation showed that the mitotic count proposed by our pipeline on a fully-automated setting produced 44.1\%, and 28.2\% less mean absolute percentage error (MAPE) compared to the baseline on the CCMCT and CMC datasets, respectively.

\section{Related Work}

To perform automatic mitosis detection, many detection algorithms have been proposed to solve this problem. Early on, hand-crafted based object detection was a popular approach for automatic mitosis detection \cite{veta2013, 6460094, 6460626, 7165640, Tek2013, 6460515, Nateghi, 10.1007/978-3-319-24571-3_12}. It was also widely used in a general computer vision tasks before the resurgence of the deep learning approach. In this approach, the object candidates were proposed first by using traditional computer vision techniques to assign the probability of each pixel being a mitotic figure, and a threshold was then applied. After that, the shape, texture, and statistical features of the mitotic figure candidates were extracted based on pathologist's knowledge. Finally, the extracted features were fed to a classifier to distinguish objects of interest from background. This approach achieved competitive performances compared to deep object detection on the ICPR MITOS-2012, AMIDA 2013, and ICPR MITOS-ATYPIA-2014 dataset. Nevertheless, this approach does not scale well to large-scale datasets, since manually designing features that could explain all the mitotic figures would be extremely labor-intensive and would not generalize well to new datasets.

Another approach for solving the problem is deep learning. It uses a convolutional neural network (CNN) to learn important features from training images. This paradigm achieves a state-of-the performance on many general computer vision tasks such as image classification, object detection, and semantic segmentation. Moreover, it could also be applied to medical imaging tasks, leading to widespread adoption \cite{LITJENS201760}. Malon et al. \cite{Malon2013} used image processing to propose the location of the candidate for mitotic cells, then used hand-crafted and CNN features to recognize mitotic figures. Cireşan et al. \cite{Cire13} trained a single-stage pixel-level classifier based on CNN to recognize mitotic figures on an image patch and perform inference in a sliding window manner, removing the need for hand-crafted features. CasNN \cite{Chen_Dou_Wang_Qin_Heng_2016} started using a two-stage pipeline to perform mitosis detection. The first stage was a semantic segmentation network trained to coarsely propose the location of mitotic cells. After that, the classification network was used to refine the prediction result observing it in fine detail. DeepMitosis \cite{LI2018121} changed the detection algorithm of the first stage from semantic segmentation to object detection, leading to a significant performance gain. In the dataset without pixel-level annotation, the bounding box was estimated using a semantic segmentation network. MitosisNet \cite{alom2019} changed the first stage by posing the problem as multi-task learning by training both segmentation and detection tasks in parallel. Though significant progress has been made, the benchmarks are mainly performed on small-scale datasets.

An introduction of large-scale mitosis detection dataset \cite{CCMCT, CMC} opened up the possibility of evaluating model performance on a whole slide level. Aubreville et al. \cite{Aubreville2020} compared three deep learning-based methods for identifying the location with the highest mitotic density in the WSI of canine cutaneous mast cell tumor. (CCMCT). It was found that a two-stage pipeline, which contains a dedicated object detector, achieved the highest correlation between the predicted and the ground truth mitotic count. In addition, the prediction proposed by the models generally performed better than individual expert. Later, Bertram et al. \cite{Bertram2021.06.04.446287} showed that the use of a model to assist  human expert by pre-selecting the region of interest led to a consistently more accurate mitotic count. In terms of speed, Fitzke et al. \cite{Fitzke2021OncoPetNetAD} proposed a high-throughput deep learning system that could perform mitosis detection on the WSI with an inference time of 0.27 minutes per slide. Most importantly, their system led to a change in tumor grading compared to human expert evaluation in some cases.

\section{Methods}

\begin{figure}[t]
\centering
\includegraphics[width=0.9\textwidth]{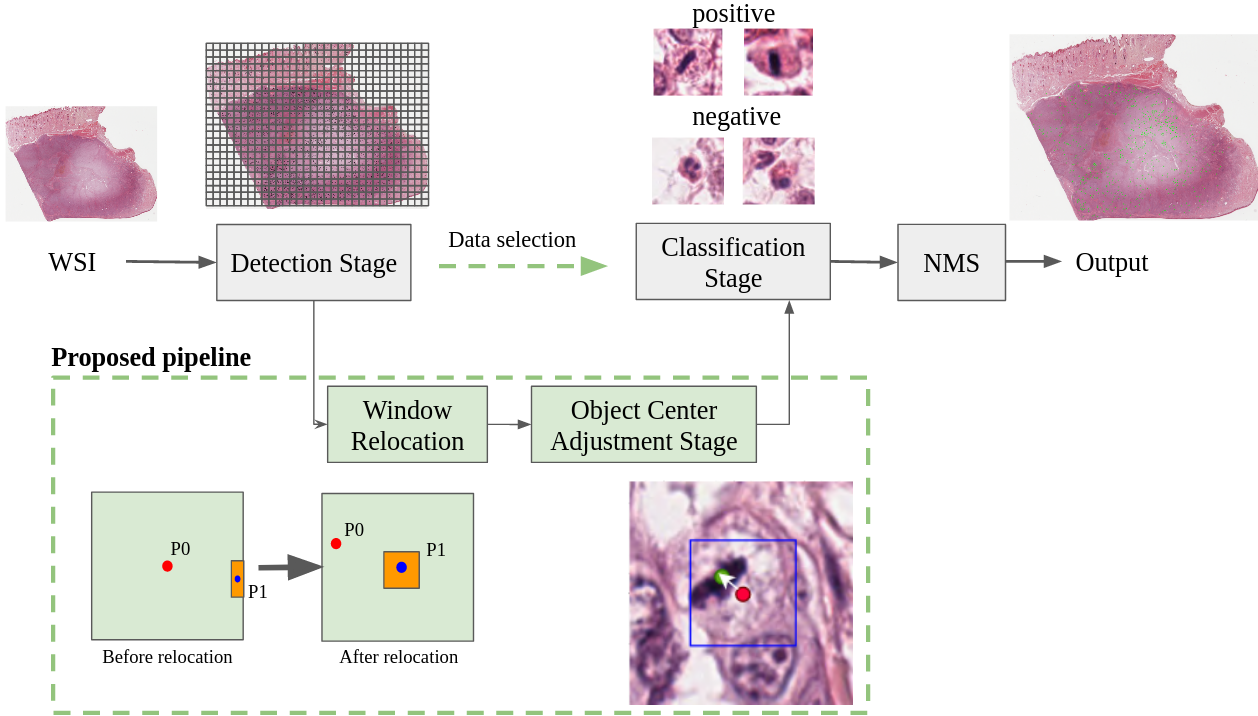}
\caption{A brief overview of our proposed pipeline. Our contributions are highlighted in the green dashed box. Two new stages, namely window relocation and object center adjustment are introduced in our pipeline. Window Relocation is used to remove superfluous poor quality predictions around the sliding window borders. The object center adjustment stage is responsible for aligning the center of the predicted positive class object from the detection stage to the image patch center. Data selection is used to filter training examples from the WSI to improve the model in the classification stage.}
\label{fig:main_pipeline}
\end{figure}

In this section, we explain each component of our proposed pipeline in full detail. An overview of our pipeline is shown in Figure \ref{fig:main_pipeline}. The pipeline consists of four stages. First, a detection stage proposes the location of the mitotic figures in the WSI using an object detection algorithm. After that, a window relocation algorithm reevaluates poor quality false positive predictions around the image border. Then, an object center adjustment stage refines the quality of the extracted object to be more aligned to the patch center. Finally, a classification stage rescores the object confidence of each patch. In the classification stage, an additional technique is used to select training examples from the WSI to boost the model performance by using disagreement between the detection and classification stage. The subsections provide a detailed explanation of each stage.  

\subsection{Detection Stage}

The detection stage is the first component of the pipeline responsible for proposing the location of the mitotic figures from the image. It is a deep object detector that receives an image as an input and return a set of bounding boxes \{($x_1$, $y_1$, $w_1$, $h_1$, $S_1$), ..., ($x_n$, $y_n$, $w_n$, $h_n$, $S_n$)\}, where each tuple in the set represents the center of the predicted object, object width, object height, an positive object confidence, respectively. Due to the sheer size of the WSI, the slide is broken down into smaller patches in a sliding window manner. The sliding window algorithm breaks down the slide with the dimension of $W \times H$ into $\lceil\frac{W}{K}\rceil \times \lceil\frac{H}{K}\rceil$ image patches (window) with the window size of $K \times K$. The detection stage then performs inference on every patch to extract the location of the mitotic figures inside it. To train the detector, we follow the data sampling strategy of the CCMCT and CMC baseline \cite{CCMCT, CMC}. To stabilize the model performance, we slightly modify the training process by sampling training images beforehand instead of querying them  on the fly.

The use of the sliding window algorithm leads to overproduced poor quality false positive predictions. This is because the object around the window boundary might be partially split into multiple objects in multiple sliding windows. Thus, an overlapping sliding window is performed to mitigate this issue by allowing the patches to be overlapped with the former one. This results in partially split boxes around the window border getting fully covered, though redundant predictions are also excessively produced. Therefore, non-maximum suppression (NMS) is used as a post-processing method to remove redundant objects. The NMS suppresses the bounding box when there exist nearby bounding boxes of which an intersection over union (IOU) is over a certain threshold and has higher confidence. The use of NMS results in a reduction of false-positive predictions as low-quality, low confidence boxes are mostly removed while retaining the good quality, high confidence ones. Despite the advantage, the overlapping windows increased the number of patches to perform inference to  $\lceil\frac{W}{K(1 - \sigma)}\rceil \times \lceil\frac{H}{K(1-\sigma)}\rceil$, where $\sigma$ is an overlapping ratio. Moreover, though the problem is mitigated, this method does not guarantee good performance at the borders.

\subsection{Window Relocation}

\begin{figure}
\centering
\includegraphics[width=0.99\textwidth]{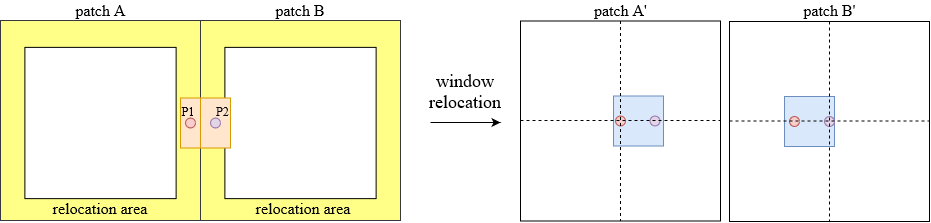}
\caption{An illustration of the window relocation algorithm. An object of interest (orange box) resides within non-overlapping sliding windows A and B. As a result, patches A and B each produce a poor-quality box whose center is the point P1, and P2 respectively. Since both centers are in the relocation area, they are valid candidates for relocating. The window relocation algorithm starts by discarding the two boxes in both patches. Then, patches A' and B' are newly created with the points P1 and P2 at the patch center. Finally, the newly created patches are fed to the detector, which returns two blue boxes.}
\label{fig:refocusing}
\end{figure}

Window Relocation is a simple algorithm used to remove poor quality predictions around the sliding window border. This method aims to eliminate the two main weaknesses of the overlapping sliding window. The first weakness is that poor quality predictions around the window border still exist when the IOU is not high enough for the NMS to suppress, which results in an increased number of false positives during the final evaluation. Another weakness is that the computation resource is wasted when the window and its surroundings do not contain any object, especially for this task where mitotic figures are often sparsely populated across the WSI. 

Figure \ref{fig:refocusing} illustrates the process of the window relocation algorithm. Window relocation mitigates both problems by performing three steps. First, a relocation area is defined around the border of each patch (the yellow area in Figure \ref{fig:refocusing}). All positive objects whose center resides in the area are then discarded.
After that, for each discarded object, the new window whose center is the center of the discarded object is created (patch A' and B' in Figure \ref{fig:refocusing}). Finally, the detector performs inference on the newly created windows. By performing these steps, the focus of the object is moved from the window border to the newly-created window center. This algorithm provides us with three advantages. First, it would reduce the poor quality predictions around the window border as most of them are removed. Second, having a relocated object positioned at the window center results in a more consistent detection result. Third, this method does not increase computation cost in the area that does not contain any object. Though this method might incur redundant predictions, it does not pose a significant impact on the whole pipeline as the new consistently produced boxes could be easily removed by using NMS.

Next, we define a clear definition of a relocation area. The $i^{th}$ object in each window could be considered to be in the relocation area if the condition below is satisfied.

\begin{equation}
( min({x_i, y_i, K - x_i, K - y_i}) \leq M ) \wedge (S_i \geq D))
\end{equation}

In other words, the center of the object that is less than equal to $M$ pixel from the window border in any axis and has higher positive object confidence than $D$ is in a relocation area and is eligible for window relocation.

$M$ is a hyperparameter determining a distance threshold from a window border, affecting the number of re-observed objects. If $M$ is set to a low value, window relocation would act as a non-overlapping sliding window. In contrast, a high value of $M$ would allow more objects to be re-scored. Setting $M$ to a high value would also come with a trade-off because it would result in an increased computation cost since the detector has to re-inference more objects.  Nevertheless, the use of window relocation is expected to have less computation costs than the overlapping sliding window. This is because it would only try to re-inference the objects around the window border, and the objects in the datasets are often sparse. $D$ is a positive confidence threshold used for discarding obvious negative objects produced by the detector. It is set to 0.05 for both datasets.

Since we know beforehand during the annotation process that the mitotic figure often has a form of circular shape with a radius around 25 pixels, we also follow this assumption and set $M$ to 25 pixels. It should be noted that this method would not work efficiently on general object detection tasks as the object shape could not be known beforehand.

\subsection{Object Center Adjustment Stage}

\begin{figure}[ht]
\centering
\includegraphics[width=0.8\textwidth]{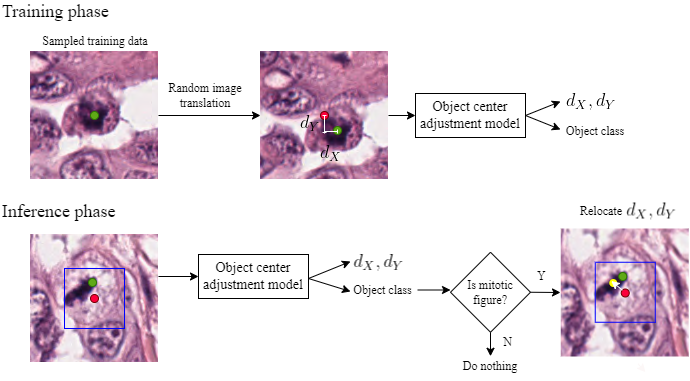}
\caption{An overview of an object center adjustment stage. The object center adjustment stage learns to estimate the distance from the extracted patch center (red dot) to the ground truth positive class object center (green dot) and its class. During inference, the model estimated the location of object center (yellow dot) and generates a new image patch at the predicted location if the predicted object is recognized as a positive class. The blue box is a bounding box predicted by the detection stage.}
\label{fig:relocation}
\end{figure}

Although many false-positive samples around the border of the sliding window are reevaluated after window relocation, there is still the problem of poor-quality bounding boxes that cause input inconsistency at the classification stage. The input inconsistency could make extracted object not being positioned at the image patch center, leading to classification stage performance degradation due to input translation variance. Therefore, we introduce an object center adjustment stage as a refinement process after window relocation to reduce position inconsistency of the positive class objects in the image patch by making the object center more aligned to the center of the patch to reduce input translation variance. The object center adjustment stage is a model which learns to locate the center of the positive object by estimating the distance from the image patch center to the ground truth positive class object center. Then, during an inference, it predicts the object center location and generates a new patch of which the center is the predicted location if the object class is positive. The negative class objects are refrained from adjustment because the concept of object center is ambiguous for non-cell background and broad tissue texture areas. Figure \ref{fig:relocation} shows an overview of the object center adjustment stage.

To train the model to estimate the position of the object center, we generate the data representing the object center at different locations in the patch as an input to the model. The generation process starts by randomly sampling positive and negative objects from the dataset and extracting them in an image patch. By doing so, the image center of the sampled object is always at the same position as the ground truth object center. Then, random geometric transformations, which are random image shifting, flipping, rotation, are applied to the sampled image. As a result, the ground truth center is shifted from the image center by $(d_X, d_Y)$ pixels. After the image is transformed, the model learns to predict the position of the object center by predicting $(d_X, d_Y)$. The value of $d_X, d_Y$ is drawn from a normal distribution and is limited to a small value ($d_X, d_Y \leq $ 12 pixels) because we assume that the center of the predicted object should be close to the ground truth object center. 

Since the objective of this stage is to relocate the center of the positive object, the class of the object has to be known beforehand, which is not practical in a real-world situation. Therefore, the object class has to be inferred from the model. We could straightforwardly obtain the class by using object confidence from the detection stage. The detected object could be inferred as a positive class when the confidence is above a certain threshold. However, using detector confidence might not be ideal as the confidence produced by the poor bounding boxes might be inaccurate. Therefore, we added an auxiliary task for the object center adjustment stage to classify the object class. Since the input to this stage is just an extracted patch, it allows the model to observe a single object at a time, removing an unnecessary distraction from other objects. As a result, the confidence produced by this improvement should be superior to the detector confidence because it inherits the advantage of the limited observation like the classification stage, and it also has information of the annotated object center.

 The object center adjustment stage is a deep convolutional neural network (CNN) that outputs two prediction heads: the main regression head to estimate the distance from the image center to the ground truth center $(d_X, d_Y)$, and the auxiliary classification head to predict the object class. The model is optimized using relocation loss $L_{rel}$ as shown below. 
 \begin{equation}
 L_{rel} = \lambda_{reg}L_{reg} +  ( 1 - \lambda_{reg})L_{cls}.
 \end{equation}

The relocation loss $L_{rel}$ is a combination of the regression loss $L_{reg}$ and classification loss $L_{cls}$ weighted by the parameter $\lambda_{reg}$. The classification loss is a standard cross-entropy loss calculated between the predicted and the ground truth object class. The regression loss is a $L1$ loss calculated between the predicted and the ground truth object center distance. To prevent a regression noise, the regression loss calculation is ignored when the ground truth class is negative. 

During inference, the model receives an extracted object as input then returns the object class and location of its center by estimating the distance from the object center to the patch center as an output. If the predicted object confidence is above a certain threshold, the object would be considered a positive object, and a new patch of which the center is the predicted location is generated. On the other hand, the model does nothing if the object's confidence is below the threshold.

\subsection{Classification Stage}

After the object center adjustment stage is performed, the center of the extracted object moves closer to the patch center and is ready to be fed to the classification stage. A classification stage is a model that resembles the object center adjustment stage but is dissimilar in its functionality. In contrast to the previous stage, this stage is a CNN that only outputs a classification head. The classification stage receives an extracted object from the object center adjustment stage as an input and returns the object's confidence. It could be argued that this stage might be redundant as the object center adjustment stage could also return the confidence. However, the main difference from the previous stage is that the object is consistently positioned at the image center. This means that the importance of having the model captured object translation variance is lessened. As a result, data augmentation strategies that could change the location of the object center are not included during training, leading to an increase in training stability and better recognition performance. 

The training process of this stage is similar to the object center adjustment model. First, positive and negative objects are randomly sampled from the dataset in an isolated area. The samples are then augmented and fed to the classifier, which predicts the object confidence. We follow DeepMitosis \cite{LI2018121} for the final object confidence calculation. The final object confidence $S$ is weighted between the confidence produced by the detection stage $S_{det}$ and the classification stage $S_{cls}$ using the weight $\omega$ as shown below. 
 
  \begin{equation}
S = \omega S_{det} +  ( 1 - \omega) S_{cls}.
 \end{equation}

\subsection{Active Learning Data Selection}

Though the proposed pipeline yields a amiable performance, the dataset is still not fully utilized. This is because the classification stage only observes annotated objects, and the unannotated ones are left untouched. DeepMitosis\cite{LI2018121} tackled this issue by using the detector to extract image regions from the original WSI to to train the classification stage. However, this method  became less effective in a large-scale dataset because it would generate an enormous number of objects from the negative class from the WSIs. Inclusion of these additional data would introduce not only a severe class imbalance but also the issue of negative class's uninformativeness. Therefore, active learning techniques should be used to select only the informative subset of proposed objects.     

To quantify the informativeness of a proposed object, we use an L1 distance between the positive class confidence of the detector and the classifier. This criterion offers us two advantages. First, it would encourage the classifier to correct its mistake by learning from the detector which generally performs better at filtering out negative objects. Second, it discourages the selection of noisy annotations, since it is possible that many  objects of the positive class were not annotated as such. In these cases, both the detector and the classifier would return high positive class confidences and discard them. Here, we select top N (N = 20,000) negative objects which has the highest informativeness as additional queries for retraining the classification model.

\section{Experimental Setup}

\subsection{Dataset}
The datasets chosen for benchmarking of our method were the ODAEL variant of the CCMCT dataset \cite{CCMCT} and the CODAEL variant of the CMC \cite{CMC} dataset. The prominent characteristic of the two datasets was the availability of a complete mitotic figure annotation on the WSI level using algorithm-aided annotation and the consensus of experts. In addition, hard negative objects (mitosis figures lookalikes) were also annotated, which improve training information. The CCMCT dataset contains an annotation of 44,800 mitotic figures on 32 WSIs, of which 11 of them were held out for testing. The CCMCT dataset consists of four classes: Mitosis, Mitosislike, Granulocyte, and Tumorcell. The first class is a positive class while the rest are considered negative. In the same manner, the CMC dataset contained an annotation of 13,907 mitotic figures on 21 WSIs, of which 7 of them were held out for testing. The CMC dataset consists of two classes: Mitosis, and Nonmitosis. 

\subsection{Detection Stage}
The training was conducted using Faster R-CNN \cite{NIPS2015_14bfa6bb} with ResNet-50 \cite{he2015deep} as a network backbone with an input training resolution of $512 \times 512$. The network backbone was initialized using ImageNet pre-trained weights  \cite{5206848}. We did not modify the base detection algorithm except for the number of output classes. We sampled 5,000 image patches from each training slide using the same data sampling strategy as the baseline. The training framework was based on an object detection framework MMDetection \cite{chen2019mmdetection}.
The model was trained using a batch size of 8 and SGD as an optimizer. The model was trained with an initial learning rate of $10^{-3}$ for 8 epochs which were divided by 10 after 5 and 7 epochs. Random flip and standard photometric augmentation were used during training.  

\subsection{Object Center Adjustment Stage}
The training was conducted using EfficientNet-B4 \cite{tan2020efficientnet} as a network backbone with an input training resolution of $128 \times 128$. The network backbone was initialized using ImageNet \cite{5206848} pre-trained weights. The model was trained using a batch size of 64 and Adam as an optimizer. The model was trained with an initial learning rate of $10^{-4}$  for 30,000 iterations which were divided by 10 after 22,500 and 27,000 iterations. $\lambda_{reg}$ was set to 0.95 for every experiment. Random image geometric and standard photometric augmentation was used during training. The positive class threshold was set to 0.2, and 0.5 for CMC and CCMCT datasets, respectively.  

\subsection{Classification Stage}

The training was conducted using EfficientNet-B4 \cite{tan2020efficientnet} as a network backbone with an input training resolution of $128 \times 128$. The network was initialized using ImageNet pre-trained weights. The model was trained using a batch size of 64 and Adam as an optimizer. For the CCMCT dataset, the model was trained with an initial learning rate of $5 \times 10^{-4}$ for 30,000 iterations which were divided by 10 after 22,500 and 27,000 iterations. For the CMC dataset without data selection, the model was trained with an initial learning rate of $5 \times 10^{-4}$ for 15,000 iterations which were divided by 10 after 10,000 and 13,000 iterations. For the CMC dataset with data selection, the model was trained with an initial learning rate of $5 \times 10^{-4}$ for 24,000 iterations which were divided by 10 after 15,000 and 21,000 iterations. Random image geometric and standard photometric augmentation except for random translation were used during training. 


\section{Results}
In this section, we evaluated the performance of the proposed method on the CCMCT \cite{CCMCT} and CMC \cite{CMC} datasets. We followed the prior study \cite{CCMCT} by using F1 (\%) as a primary metric and using the same train-test split. We reported an average of three splits with standard deviations. The models used for evaluation were the checkpoints at the last training step.

The result shown in table \ref{mainresult} summarized the performance of our method. Ultimately, the performance of the proposed pipeline improved from 82.0\% to 83.2\% on the CCMCT dataset and 77.5\% to 82.6\% on the CMC dataset. The main contributing factors were data selection and object center adjustment stage, which contributed 2.6\% and 4.2\% absolute performance improvement. The result suggested that input consistency and exposure of additional unannotated data at the classification stage was crucial for performance improvement. 

\begin{figure}
     \centering
     \begin{subfigure}{0.45\textwidth}
         \centering
         \includegraphics[width=\textwidth]{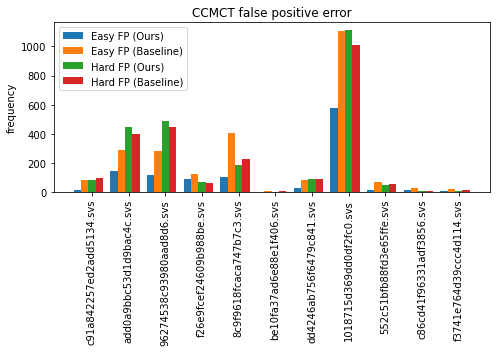}
         \caption{}
     \end{subfigure}
     \begin{subfigure}{0.45\textwidth}
         \centering
         \includegraphics[width=\textwidth]{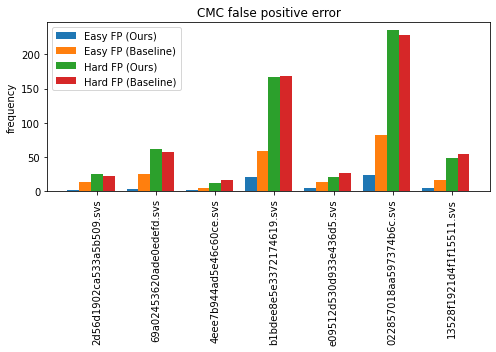}
         \caption{}
     \end{subfigure}
    \caption{Multiple Bar charts showing the frequency of easy and hard false positive (FP) errors on the CCMCT and CMC dataset. Our method greatly reduced the number of easy false positive predictions, yet confusion between positive and hard-negative samples still remained in high quantity.}
    \label{fig:error_result}
\end{figure}

We then investigated the mispredictions produced by our pipeline by observing false-positive errors and categorized them as easy and hard errors. The hard errors are the hard-negative object that is confused as a positive class, while the easy error is confusion between the positive class and a non-hard negative object or background image. Figure \ref{fig:error_result} shows a visualization of false-positive errors of our method. Our method greatly reduced the number of easy false positive predictions compared to the baseline. Nevertheless, the confusion between positive and hard-negative samples persists. This indicated that input translation variance was not the only factor for the confusion between hard-negative and positive objects.
\begin{table}
\caption{The test F1 (\%) performance of the proposed method evaluated on the CCMCT and CMC datasets. $\pm$ denotes standard deviation.}
\begin{center}
\begin{tabular}{|l|c|c|}
\hline
\textbf{Method}       & \textbf{CCMCT test F1(\%)} & \textbf{CMC test F1(\%)}  \\ \hline
Baseline \cite{CCMCT, CMC} & 82.0  & 77.5 \footnotemark\\ \hline
Reproduced baseline ($\omega = 0$)                  & 79.9 $\pm$ 0.3 & 77.6 $\pm$ 0.2 \\ \hline
+ Data selection  & 81.8 $\pm$ 0.1 & 80.3 $\pm$ 0.1\\  \hline
+ Object center adjustment          & 82.5 $\pm$ 0.1 & 81.8 $\pm$ 0.1\\  \hline
+  Weighted confidence ($\omega = 0.4$)    & 83.0 $\pm$ 0.1 & 82.1 $\pm$ 0.1\\ \hline
+ Window relocation     & \textbf{83.2 $\pm$ 0.1} & \textbf{82.3 $\pm$ 0.1}\\ \hline
\end{tabular}

\label{mainresult}
\end{center}
\end{table}
\footnotetext{The number was based on the erratum in their Github.}

\subsection{Effect of Object Center Adjustment Stage}

In this subsection, we study the effect of the object center adjustment stage on the proposed pipeline. First, we show that the presence of this stage leads to an improvement of the proposed object center quality. Then, we provide ablation studies to confirm the choice of our design. For every experiment, $\omega$ was set to zero, and window relocation was excluded.

One metric that can measure the performance of the object center adjustment stage is the distance between the patch center and the original location. The false positives were not included in this metric as it was irrelevant for this stage. The object center adjustment stage reduced the average distance from 3.59 to 3.17 on the CCMCT dataset and 3.61 to 3.40 on the CMC dataset. The result suggested that the use of the object center adjustment stage clearly reduces the input translation variance.

\begin{figure}
     \centering
     \begin{subfigure}{0.8\textwidth}
         \centering
         \includegraphics[width=\textwidth]{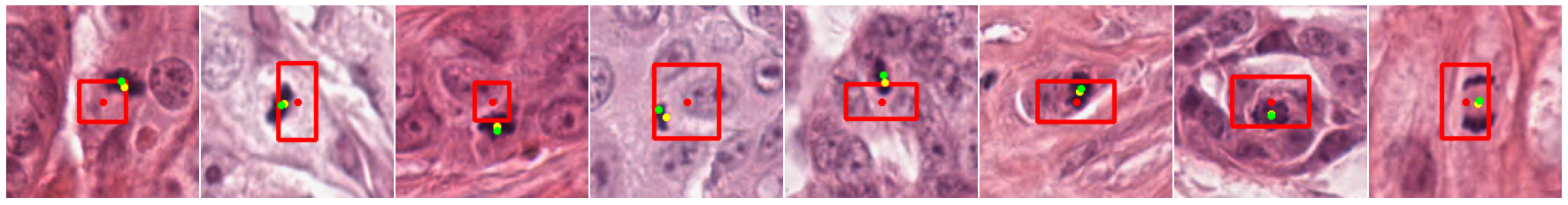}
         \caption{Successful cases of object center adjustment stage.}
         \label{rel_success}
     \end{subfigure}
     \begin{subfigure}{0.8\textwidth}
         \centering
         \includegraphics[width=\textwidth]{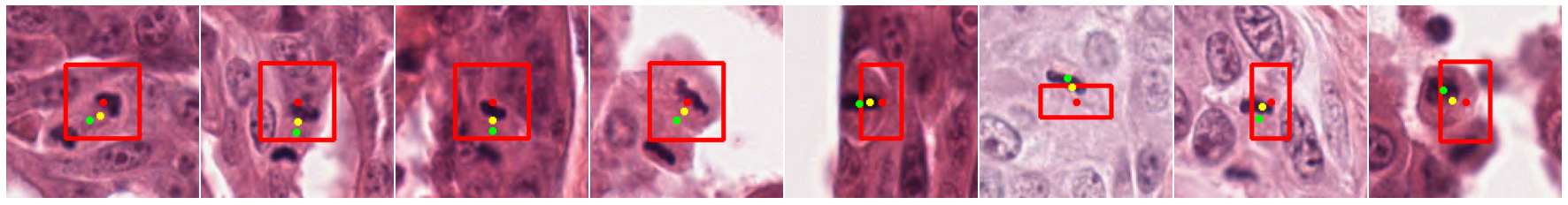}
         \caption{Failure cases of object center adjustment stage.}
         \label{rel_fail}
     \end{subfigure}
    \caption{Example prediction results produced by the object center adjustment stage on the CMC dataset. (a) shows a successful examples. (b) shows failure examples. The first four images of are failures at the telophase stage. Red, yellow, and green dots indicate original, relocated, and ground truth object center, respectively. The red boxes are the bounding box produced by the detection stage. }
    \label{fig:qualitative_relocation}
\end{figure}

Figure \ref{fig:qualitative_relocation} shows examples of the predicted object center produced by the object center adjustment stage. The model often correctly located the position of the actual object center as shown in Figure \ref{rel_success}. However, falsely adjusted objects were also present. Some common mispredictions came from confusion of cells in the late telophase stage which can look like two separate mitotic figures. As a result, the model aligned to one of the spindles instead of the actual center. Others causes of misprediction came from the model's inability to precisely locate the object center when the predicted object center is too far from the ground truth center, object center ambiguity, and silly mistakes.

Next, we justify the exclusion of negative class in the regression loss and the presence of auxiliary head. Table \ref{relocation_ablation} shows that the model performance reduced from 81.8\% test F1 to 81.1\% when the negative class was included in the regression loss. The result indicates that the ambiguity of object center in the negative class object led to a regression noise during training, eventually leading to reduced performance. Moreover, the auxiliary head improves the model performance from 81.5\% to 81.8\%, showing the importance of multi-task learning.


We also conducted ablation studies on the choice of pipeline design and the removal of data augmentation strategies that could change the location of the object center. Table \ref{relocation_importance} shows that translation augmentation improved the performance of the classification stage of the base pipeline. However, the object center adjustment training scheme, which formulated the problem as a multi-task problem, is more efficient than data augmentation. We confirmed this by replacing a classification stage with an object center adjustment stage and using its classification head to produce object confidence. It was found that, by only using the object center adjustment stage, the performance of the whole pipeline improved from 80.5\% to 81.3\% test F1 on the CMC dataset. By stacking the relocation and classification stage, the performance was further increased to 81.8\%. However, having a translation augmentation in the classification stage of the stacked pipeline degraded the performance. The result also indicated that translation augmentation hampered the performance when the translation variance of the object center was controlled. 

\begin{table}
\caption{The result of the ablation study on the importance of the object center adjustment stage conducted on the CMC dataset. The use of an object center adjustment stage outperformed the classification stage with translation augmentation. In addition, the removal of translation augmentation at the classifications stage was crucial for the performance improvement of the whole pipeline.}
\begin{center}
\begin{tabular}{|l|c|c|c|}
\hline
\textbf{Method}  &  \textbf{CMC test F1(\%)}  \\ \hline
 Classification stage  & 80.3 $\pm$ 0.1 \\ \hline
 Classification stage w/ translation augmentation   & 80.5 $\pm$ 0.3 \\ \hline
 Object center adjustment stage   &  81.3 $\pm$ 0.1 \\ \hline
 Object center adjustment stage+Classification stage &  \textbf{81.8 $\pm$ 0.1} \\ \hline
 Object center adjustment stage+Classification stage w/ translation augmentation &  81.5 $\pm$ 0.1 \\ \hline

\end{tabular}

\label{relocation_importance}
\end{center}
\end{table}

\begin{table}
\caption{The result of the ablation study of the object center adjustment stage conducted on the CMC dataset. The use of an auxiliary head improved the stage performance while the inclusion of negative class for relocation loss resulted in reduced performance.}
\begin{center}
\begin{tabular}{|c|c|c|c|}
\hline
\textbf{Negative class relocation loss} & \textbf{Auxiliary head} &  \textbf{CMC test F1(\%)}  \\ \hline
 - & -  & 81.5 $\pm$ 0.2 \\ \hline
 \checkmark & - & 81.1 $\pm$ 0.1 \\ \hline
 - & \checkmark  & \textbf{81.8 $\pm$ 0.1} \\ \hline
 
\end{tabular}

\label{relocation_ablation}
\end{center}
\end{table}

\subsection{Effect of Window Relocation}
This subsection aimed to measure the effect of window relocation on the whole pipeline. Table \ref{refocusing_ablation} shows a comparison between window relocation and the sliding window method. The use of overlapping sliding windows did not improve the performance of our pipeline as most of the overproduced samples could be removed using the object center adjustment stage and non-maximum suppression. By using window relocation, the performance of the pipeline was better than the non-overlapping sliding window and the overlapped one with 0.2\% test F1 absolute improvement on the CMC dataset. The result suggested that some produced errors could not be mitigated through the method above. This is because the center of the overproduced object might be too far for the object center adjustment stage to adjust back to the actual center. In addition, we also found that window relocation only incurs a small amount of additional inference time over non-overlapping sliding window in a practical setting. This is because mitotic figures in the WSI generally have low density. Moreover, unlike overlapping sliding windows, window relocation could ignore most of the background image as it did not contain any objects in the first place.

Since both window relocation and object center adjustment stage have a similar objective of improving poor quality predictions for the detection stage, we conducted an ablation study to observe the effect of each component separately. Table \ref{refocusing_ablation2} shows a comparison of the two components on the CMC dataset. Window relocation improved the test F1 from 80.3\% to 81.1\%. Nevertheless, the performance was inferior to the object center adjustment stage, which achieved 81.8\%. This is because window relocation mostly affects the objects positioned around the sliding window border. 

\begin{table}
\caption{A comparison of different sliding window algorithms on the CMC dataset. Window relocation outperformed overlapping sliding windows while incurring less computation cost. }
\begin{center}
\begin{tabular}{|l|c|c|}
\hline
\textbf{Method} & \textbf{CMC test F1(\%)} &  \textbf{\thead{Number of test inference window\\ done in the detection stage}  }\\ \hline
 Non-overlapping sliding window &  82.1  $\pm$ 0.1 & 211482 (+0\%)\\ \hline
 Overlapping sliding window &  82.1  $\pm$ 0.1 & 261909 (+23.8\%) \\ \hline
 Window relocation &  \textbf{82.3  $\pm$ 0.1} & 217368 (+2.7\%) \\ \hline

\end{tabular}

\label{refocusing_ablation}
\end{center}
\end{table}

\begin{table}
\caption{The comparison between window relocation and object center adjustment stage on the CMC dataset. Window relocation could partially mitigate the problem of input translation variance.}
\begin{center}
\begin{tabular}{|c|c|c|}
\hline
\textbf{Window relocation} & \textbf{Object center adjustment stage}  & \textbf{CMC test F1(\%)}  \\ \hline
 - & - &  80.3 $\pm$ 0.1 \\ \hline
 \checkmark & -  & 81.1 $\pm$ 0.2 \\ \hline
 - & \checkmark  & 81.8 $\pm$ 0.1 \\ \hline
 
  \checkmark & \checkmark  & \textbf{82.1 $\pm$ 0.1} \\  \hline

\end{tabular}

\label{refocusing_ablation2}
\end{center}
\end{table}

\subsection{The effect of the detection algorithm}
We conducted an ablation study on the detection algorithm on the CMC dataset by changing the base detection algorithm and found that our method reduced the dependence on the strength of the detection model. We compared the chosen detection algorithm, Faster-RCNN-ResNet50, to the RetinaNet-ResNet18 \cite{retina}, which is a detection algorithm in both CCMCT and CMC paper. We also observed the effect of different model backbones by comparing them with Faster-RCNN-ResNet101. Every experiment was trained using the same set of data and training schedule, and $\omega$ was set to 0. Table \ref{detection_ablation} shows that the choice of detection algorithm had a significant impact on the base pipeline. The use of RetinaNet-ResNet18 as a detection algorithm reduced the test F1 on the CMC dataset by 9.0\% on the detection stage and 2.2\% on the classification stage compared to Faster-RCNN-ResNet50. The presence of object center adjustment stage and data selection helped mend the performance gap from 2.2\% to 0.5\% test F1 difference. The result suggested that the performance of the detection algorithm has a direct impact on the quality of the predicted bounding box, leading to worse classification stage performance, and the object center adjustment stage is an essential component for the object center correction. It could also be implied that, by emphasizing the classification stage, it allowed us to use a fast detection algorithm to potentially greatly reduce the inference time of the detection stage while not suffering a sharp performance reduction. 
 \begin{table}
\caption{A comparison of different detection algorithms on the CMC dataset. The choice of different detection algorithms led to drastically different results on the classification stage. However, an introduction of the object center adjustment stage and data selection greatly mitigated this problem.}
\begin{center}
\begin{tabular}{|c|c|c|c|c|}
\hline
\textbf{Detection algorithm} & \textbf{\thead{detection \\ test F1(\%)} } &  \textbf{\thead{+classification \\ stage}} & \textbf{+data selection} & \textbf{\thead{+object center \\ adjustment}} \\ \hline
 RetinaNet-ResNet18 &  61.4 $\pm$ 0.8  & 75.4 $\pm$ 0.2 & 78.8 $\pm$ 0.4 & 81.3 $\pm$ 0.2 \\ \hline
 Faster-RCNN-ResNet50 &  70.4 $\pm$ 0.3 & 77.6 $\pm$ 0.2 & 80.3 $\pm$ 0.1 & 81.8  $\pm$ 0.1 \\ \hline
 Faster-RCNN-ResNet101 &  71.3 $\pm$ 0.1 & 78.1 $\pm$ 0.2 &  80.2 $\pm$ 0.1 &  81.8 $\pm$ 0.1 \\ \hline

\end{tabular}

\label{detection_ablation}
\end{center}
\end{table}

\subsection{End-to-End evaluation}

\begin{figure}
     \centering
     \begin{subfigure}{0.3\textwidth}
         \includegraphics[width=\textwidth]{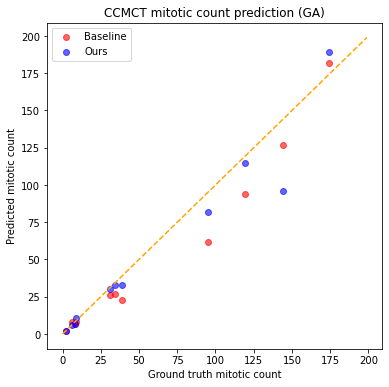}
     \end{subfigure}
     \begin{subfigure}{0.3\textwidth}
         \includegraphics[width=\textwidth]{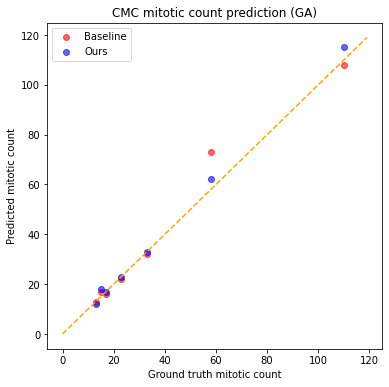}
     \end{subfigure}
    \centering
    \caption{Scatter plots illustrating the predicted mitotic count and the ground truth on the CCMCT and CMC dataset on GA setting. Compared to the baseline, our method clearly changed the predicted MC when the object appeared in high density, though the effect become less noticeable on the slides with low mitotic figures.}
    \label{end2end0}
\end{figure}

We further extended an evaluation of our method to an end-to-end setting by comparing the mitotic count (MC) produced by our method to the ground truth mitotic count. We follow Meuten et al.\cite{meuten} by counting mitotic figures at 10 HPF (2.37 $mm^2$) with an aspect ratio of 4:3 at the area with the highest mitotic figures density. The HPF area was calculated by selecting the rectangle window size of 7110/5333 pixels which contains the highest number of mitotic figures \cite{Bertram2019ComputerizedCO}. We evaluated the proposed pipeline on two settings: GA, and GB. The GA setting directly compared the mitotic count from the HPF proposed by our pipeline to the ground truth mitotic count. In contrast, the GB setting only used the proposed HPF, but the mitotic count was instead obtained by counting the ground truth mitotic cell. The GA setting could be considered as a fully automated mitosis counting while the GB was a human-in-the-loop setting where the optimal pathologist, who always correctly recognized mitotic figures, was also included in the pipeline. Moreover, the GB setting put an importance on the quality of the proposed HPF over the predicted mitotic count, which was mainly focused on the GA setting. We reported mean absolute percentage error (MAPE) and mean absolute error (MAE) at the prediction threshold which yielded the lowest MAPE. For a baseline comparison, we used the prediction results on the test set in their GitHub. Table \ref{end2end1} shows the result of GA and GB settings on the CCMCT and CMC dataset. Our method significantly reduced the MAPE and MAE on the CCMCT and CMC datasets in both settings. Figure \ref{end2end0} shows a relation between the predicted mitotic count and the ground truth. Compared to the baseline, our method clearly changed the mitotic count when the object appeared in high density, though the impact was lessened in a low-density case.

\begin{table}[htbp]
\caption{The end-to-end performance of the proposed method evaluated on the CCMCT and CMC datasets. Our method consistently outperformed the baseline in both settings.}
\begin{center}

\begin{tabular}{|c|l|cc|cc|}
\hline
\multicolumn{1}{|l|}{\textbf{Dataset}} & \textbf{Method} & \multicolumn{2}{c|}{\textbf{GA}} & \multicolumn{2}{c|}{\textbf{GB}}                                       \\ \cline{3-6} 
\multicolumn{1}{|l|}{}                                  &                                  & \multicolumn{1}{l|}{\textbf{MAPE}} & \multicolumn{1}{l|}{\textbf{MAE}} & \multicolumn{1}{l|}{\textbf{MAPE}} & \multicolumn{1}{l|}{\textbf{MAE}} \\ \hline
CCMCT  & Baseline & \multicolumn{1}{c|}{18.8} & 10.5 & \multicolumn{1}{c|}{11.2} & 4.4  \\ \cline{2-6} 
& Ours & \multicolumn{1}{c|}{\textbf{10.5}}  & \textbf{8.3} & \multicolumn{1}{c|}{\textbf{6.8}}  & \textbf{1.9} \\ \hline
CMC   & Baseline  & \multicolumn{1}{c|}{7.8} & 3.1 & \multicolumn{1}{c|}{8.1} & 2.4 \\ \cline{2-6} 
& Ours & \multicolumn{1}{c|}{\textbf{5.6}}  & \textbf{1.9} & \multicolumn{1}{c|}{\textbf{5.6}}  & \textbf{1.6} \\ \hline
\end{tabular}
\label{end2end1}
\end{center}
\end{table}

\section{Conclusion}
We propose ReCasNet, an enhanced deep learning pipeline that introduces three improvements to the two-stage mitosis detection pipeline. First, we introduced window relocation, a method used to reduce the number of false positives introduced by the sliding window algorithm by removing predictions around the window border and assigning them to a new window for re-performing inference. Second, we proposed the object center adjustment stage, a deep learning model responsible for adjusting the center of the mitotic cell predicted from the detection stage. This improves the consistency of inputs for the classification stage. Third, we utilized an active learning technique to alleviate the inconsistency in training data by identifying additional informative examples, based on the disagreement between the two stages, to train the classification stage. Our proposed method significantly increases the performance of the whole pipeline on both detection of individual mitotic figures and end-to-end region-of-interest proposal and mitotic count predictions on the CCMCT and CMC dataset.

\section*{Acknowledgment}
This work was supported by the Thailand Program Management Unit (PMU-B) Grant for Multi-Institutional AI Development in Digital Pathology (to S.Sa., S.Sh., and S.Sr.) and the Grant for Supporting Research Unit, Ratchadapisek Sompoch Endowment Fund, Chulalongkorn University (to C.P., S.Sr., and E.C.).

\section*{Code availability}
All used code in the experiments is available at https://github.com/cmb-chula/ReCasNet. The implementation is based on MMDetection\cite{chen2019mmdetection} and tensorflow \cite{tensorflow2015-whitepaper}.

 \bibliographystyle{unsrtnat}
 \bibliography{ms}

\newpage
\appendix
\renewcommand \thepart{}
\renewcommand \partname{}
\part{Appendix} 
\section{Additional training details}
\subsection{Detection stage data preparation strategy}
We followed the data preparation strategy of the CCMCT and CMC baseline. 50\% of the cropped patches were randomly acquired from the training slide. 40\% were sampled to contained at least one mitotic figure in the cropped image. 10\% contained at least one mitotic figure-lookalike in the cropped image (class MitosisLike and NonMitosis in the CCMCT, and CMC dataset respectively). 

\subsection{Data augmentation strategies}
Table \ref{auglist2} shows a detailed list of augmentation strategies of the object center adjustment and classification stages. Random rotation was still allowed in the classification stage because the relocated object center patch can still rotate.

\begin{table}[b]
\caption{List of augmentation strategy of the the classification and object center adjustment stage.}
\begin{center}

\begin{tabular}{|l|c|c|c|}
\hline
\textbf{Augmentation strategy} & \textbf{Classification stage} & \thead{Object center \\ adjustment stage} & \textbf{Intensity} \\ \cline{2-3}
                                                 & \multicolumn{2}{c|}{\textbf{probability}}                 &                                     \\ \hline
Random flip                                      & 0.5                           & 0.5                       & -                                   \\ \hline
Random brightness                                & 0.5                           & 0.5                       & (0.8, 1.2)                          \\ \hline
Random contrast                                  & 0.5                           & 0.5                       & (0.8, 1.2)                          \\ \hline
Random gaussian blur                             & 0.25                          & 0.25                      & (3,3) and (5, 5) kernel             \\ \hline
Random hue                                       & 1                             & 1                         & (-0.1, 0.1)                         \\ \hline
Random rotation                                  & 1                             & 1                         & (-90, 90)                           \\ \hline
Random translation                               & 0                             & 1                         & $d_x, d_y \sim N(0, 6^2) $                \\ \hline
\end{tabular}
\label{auglist2}

\end{center}
\end{table}

\section{Additional ablation studies}
\subsection{Effect of relocation loss weight}
We investigated the effect of $\lambda_{reg}$, a hyperparameter determining an importance of the regression task in the relocation loss on the performance of the whole pipeline. We compared the effect by using different sets of $\lambda_{reg}$ and measured the performance of the whole pipeline. Table \ref{lambda_reg} shows that the performance does not vary much when $\lambda_{reg}$ is set at the high value. However, there was a degradation of the object center adjustment stage's ability to locate object center when the value of $\lambda_{reg}$ was below a certain threshold. 

\begin{table}
\caption{The of $\lambda_{reg}$ on the performance of the pipeline on the CMC dataset.}
\begin{center}
\begin{tabular}{|c|c|c|c|}
\hline
$\lambda_{reg}$  &  \textbf{CMC test F1(\%)}  \\ \hline
1  & 81.5 $\pm$ 0.2 \\ \hline
0.99  & 81.7 $\pm$ 0.1 \\ \hline
0.95   & 81.8 $\pm$ 0.1 \\ \hline
0.9  &  \textbf{81.8 $\pm$ 0.1} \\ \hline
0.8 &  81.8 $\pm$ 0.1 \\ \hline
0.7 &  81.6 $\pm$ 0.4 \\ \hline
0.6 &  80.3 $\pm$ 1.6 \\ \hline

\end{tabular}

\label{lambda_reg}
\end{center}
\end{table}

\subsection{Effect of data selection algorithm for the classification stage}

In this section, we show that our criterion of informativeness is effective for this task. Thus, we provided a comparison of our method against three baselines. The first baseline is DeepMitosis \cite{LI2018121} query strategy, which queries every negative object proposed by the classification stage from the training slides. The second baseline is uncertainty sampling, a strong baseline in the Active Learning field \cite{Settles2009ActiveLL}. This method measures the uncertainty produced by the model as a selection criterion for data acquisition. We used entropy as an uncertainty measurement and used classification stage confidence to produce model uncertainty. The third baseline is K-Center-greedy \cite{sener2018active}, a query strategy based on the core set approach. It aims to select the samples that provide the most coverage over the training distribution by minimizing the distance between a data point and its nearest chosen samples. We also follow their work by using the output after the last convolutional layer of the classification stage to represent the data point and L2 as a distance function. We used the same classification model for data acquisition for every baseline. Window Relocation and object center adjustment stage were excluded during the experiments.

Table \ref{query_strategy} shows the result of our experiment. It was found that our method outperformed DeepMitosis's querying strategy and Active Learning baselines, and every Active Learning baseline is better than not selecting any data at all. The result supported our claim that overexposure of negative samples led to sub-optimal performance but still better than not querying any additional data at all.

\begin{table}
\caption{The effect of data selection algorithm on the performance of the pipeline on the CMC dataset.}
\begin{center}
\begin{tabular}{|l|c|c|c|}
\hline
\textbf{Query method}  &  \textbf{CMC test F1(\%)}  \\ \hline
Baseline (no query)   &  77.6 $\pm$ 0.2 \\ \hline
DeepMitosis (query all)  &  80.0 $\pm$ 0.2 \\ \hline
K-Center greedy &  79.0 $\pm$ 0.1 \\ \hline
Uncertainty sampling &  79.8 $\pm$ 0.1 \\ \hline
Disagreement (Ours) &  \textbf{80.3 $\pm$ 0.1} \\ \hline

\end{tabular}

\label{query_strategy}
\end{center}
\end{table}


\end{document}